\documentclass[letterpaper, 10 pt, conference]{ieeeconf}  

\IEEEoverridecommandlockouts                              

\usepackage{cite}
\usepackage{amsmath,amssymb,amsfonts}
\usepackage{graphicx}
\usepackage{textcomp}
\def\BibTeX{{\rm B\kern-.05em{\sc i\kern-.025em b}\kern-.08em
    T\kern-.1667em\lower.7ex\hbox{E}\kern-.125emX}}
\usepackage{dsfont}
\usepackage{paralist}
\usepackage[font=footnotesize]{subcaption}
\usepackage{cite}
\usepackage{tabularx}
\usepackage{cancel}
\usepackage{bm}
\usepackage{url}
\usepackage{mathtools}
\usepackage{textcomp}
\usepackage{nccmath}
\usepackage{xcolor}
\usepackage{multirow}
\usepackage{hhline}
\usepackage{algorithm,algpseudocode}
\makeatletter
\renewcommand{\ALG@beginalgorithmic}{\small}
\makeatother
\usepackage{booktabs} 
\usepackage{makecell} 
\usepackage{xcolor}
\definecolor{myhighlight}{RGB}{255, 255, 200}

\usepackage{cleveref}

\makeatletter
\newcommand\fs@betterruled{%
  \def\@fs@cfont{\bfseries}\let\@fs@capt\floatc@ruled
  \def\@fs@pre{\vspace*{6pt}\hrule height.8pt depth0pt \kern2pt}%
  \def\@fs@post{\kern2pt\hrule\relax}%
  \def\@fs@mid{\kern2pt\hrule\kern2pt}%
  \let\@fs@iftopcapt\iftrue}
\floatstyle{betterruled}
\restylefloat{algorithm}
\makeatother

\algrenewcommand\algorithmicrequire{\textbf{Input:}}
\algrenewcommand\algorithmicensure{\textbf{Output:}}

\title{\LARGE \bf
Diffusion for Long-Horizon Multi-Robot Path Planning in Human-Shared Environments
}

\graphicspath{/Pictures}

\author{Vaibhav Sanjay$^{1}$, Yorai Shaoul$^{1}$, and Jiaoyang Li$^{1}$
\thanks{$^{1}$Vaibhav Sanjay, Yorai Shaoul, and Jiaoyang Li are with the Robotics Institute, 
        Carnegie Mellon University.
        {\tt\small \{vsanjay,yshaoul,jiaoyanl\}@andrew.cmu.edu}}%
}

\begin{document}

\maketitle
\thispagestyle{empty}
\pagestyle{empty}

\begin{abstract}
Multi-robot path planning in human-shared environments requires a delicate balance between robust inter-robot coordination and socially aware behavior. While diffusion models excel at generating predictable, human-like paths, existing generative planners are often restricted to paths of fixed duration and high computational latency, limiting their adaptability to varying goal distances and hindering real-time deployment. We present Multi-Robot Rolling Diffusion (MRRD), a novel framework that enables real-time, long-horizon navigation for large robot teams through dense crowds. MRRD combines a rolling-horizon scheme to accommodate the limited prediction horizon of human motion, parallelized diffusion inference for scalable generation of human-like paths, and a conflict-based-search mechanism for resolving inter-robot collisions. It further incorporates urgency-based temporal conditioning to generate paths with varying speeds and employs differentiated guidance terms to maximize both social awareness around humans and efficient coordination between robots. Experimental results in crowded environments demonstrate that MRRD successfully scales to 15 robots in real-time, significantly outperforming existing baselines in both safety and mission success rates.

\end{abstract}

\section{INTRODUCTION}

Multi-robot autonomous systems are successfully used in scenarios like warehouse automation, drone fleets, and inspection systems. However, their use in human-shared environments remains largely underexplored. This requires ensuring not only collision-free coordination among robots but also safe interaction with surrounding humans.

Single-robot navigation in human-shared environments has been explored extensively \cite{chenghumanshared,hanmpc,samavimpc,wangmpc,longhumanshared,Munreinforcement,mavrogiannisRL,liureinforcement,songvlm,narasvlm,wang2024sociallyadaptivepathplanning,wang2023navistar}, with a wide range of methods including human motion prediction, control barrier functions, and vision-language models. In this work, we focus on methods that directly generate human-like paths because they enable robots to move in a predictable and socially acceptable manner. Diffusion models~\cite{carvalho2023mpd}, which excel at learning from demonstrations, are therefore a natural choice for this setting. Recently, CoBL-Diffusion \cite{MizutaLeung2024} trains diffusion models on large sets of human walking data and demonstrates effective navigation for single robots in human-shared environments. However, these generative planners are often subject to a ``fixed-time problem," where they are restricted to produce paths of fixed duration regardless of the actual distance to the goal, leading to erratic velocities and reduced navigation efficiency.

Research on multi-robot path planning in human-shared environments is not as extensive and can be largely summarized in a few distinct approaches. Prioritized planning methods \cite{BajcsyScalable,KnepperSampling} anticipate human motions and plan robot paths in a predefined sequence, avoiding humans and higher-priority robots. Although quite fast, planning in sequence may result in uncooperative behavior since high-priority robots do not take the paths of low-priority robots into account. Hierarchical methods \cite{farrelNCBF,BOLDRER2022103979} plan global paths ignoring collisions and resolve them during execution by maneuvering robots around each other. Since collisions are resolved by local planning, these methods may lack the ability to perform long-term planning while maintaining social awareness. RL-based methods \cite{Wang2023RL} can directly learn robot coordination policies. However, they often rely on carefully engineered reward functions, which may not capture the full range of desirable human-like behaviors. 



This paper introduces Multi-Robot Rolling Diffusion (MRRD), a rolling-horizon approach utilizing diffusion models to generate real-time, human-like, and collision-free paths for a team of robots in human-shared environments. Rather than generating a single, globally ``pinned" path of rigid duration, our diffusion model produces localized path segments conditioned on a diverse range of urgency values. This gives MRRD significant freedom in planning for complex, dynamic scenes; it can adaptively scale velocities to suit immediate crowd density and distance to the goal.
Inspired by \cite{shaoul2025multirobot}, MRRD scales single-robot diffusion models trained from human walking data to a multi-robot setup via Conflict-Based Search (CBS)~\cite{SHARONCBS}, a classical multi-agent path-finding technique. This way, 
MRRD distinguishes between robot-robot interactions (handled by CBS) and robot-human interactions (handled by the diffusion model). We demonstrate its effectiveness in various simulated environments, showing real-time navigation with up to 15 robots in empty environments and 10 in cluttered ones.

\section{Related Works}
We briefly review related work on multi-robot path planning and robot navigation in human-shared environments.
\subsection{Multi-Robot Path Planning (Without Humans)}
Multi-robot path planning is a challenging problem of planning collision-free paths for a team of robots in a shared environment. 
One promising approach is to abstract the problem as Multi-Agent Path Finding (MAPF) \cite{Stern2019MultiAgentPD}, which assumes that robots move in discrete timesteps on a graph, typically a 2D grid. Solutions to MAPF range from optimal search-based methods to efficient heuristics. Prioritized Planning (PP) \cite{cooppath} is a fast, suboptimal algorithm that assigns fixed priorities to robots, plans their paths sequentially, and treats higher-priority robots as moving obstacles. However, PP may force lower-priority robots into circuitous routes due to sequential dependencies. The optimal and scalable standard is Conflict-Based Search (CBS) \cite{SHARONCBS}, a two-level algorithm that combines single-robot path planning with a high-level constraint tree to resolve collisions, imposing constraints only where collisions occur. Multi-Agent Reinforcement Learning (MARL) \cite{Sartoretti2018PRIMALPV} has emerged as a prominent decentralized alternative, where robots learn local navigation policies—often via actor-critic architectures.

While MAPF algorithms were originally designed for discrete settings, many have been extended to continuous time and space \cite{conttimecbs,liang2024multiagentpathfindingcontinuous}. In particular, Multi-Robot Multi-Model Planning Diffusion (MMD) \cite{shaoul2025multirobot} combines CBS with diffusion-based planners. By training on large datasets of successful single-robot paths, diffusion-based planners can generate paths that effectively capture the complex spatial dependencies of the environment and movement patterns in the data. MMD uses diffusion models to generate single-robot paths while borrowing the constraint tree structure from CBS to designate constraints to the diffusion-based planners to resolve collisions. It defines spherical constraints that push diffusion-generated paths out of collision regions without over-constraining the entire path like PP would. 

\subsection{Robot Navigation in Human-Shared Environments}
Single-robot path planning in human-shared environments has been extensively studied, utilizing approaches such as A* search \cite{chenghumanshared}, 
model predictive control \cite{hanmpc,samavimpc, wangmpc}, reinforcement learning \cite{longhumanshared, Munreinforcement, liureinforcement,mavrogiannisRL}, vision language models \cite{songvlm,narasvlm}, and generative adversarial networks \cite{wang2024sociallyadaptivepathplanning}. We are mainly concerned with algorithms that learn from human data or input, directly embedding socially aware navigation in the planning process. Navi-STAR \cite{wang2023navistar} uses preference learning with human oversight for training, but gathering preferences can be costly. CoBL-Diffusion \cite{MizutaLeung2024} learns from human walking data using a diffusion model to emulate movement and avoid paths via guidance; however, it requires knowing the full human paths beforehand, which is unrealistic.

Multi-robot navigation in human-shared environments is less studied and requires balancing high-level coordination with socially acceptable local reactivity. Existing methods can be divided into three categories: PP, hierarchical planning, and MARL. PP methods employ prioritized planning with A* \cite{BajcsyScalable} or model predictive control \cite{KnepperSampling} to avoid predicted paths of humans and higher-priority robots. PP limitations persist and can be worsened by humans cutting off paths. Hierarchical methods generate global paths without considering humans or other robots, relying on a local planner for collision avoidance during execution. Recent work integrates neural control barrier functions for safety guarantees \cite{farrelNCBF} or uses a Lloyd-based planner with Voronoi-partitioning to shift paths away from others and predicted human zones \cite{BOLDRER2022103979}. These controllers lack the long-horizon foresight needed for complex social dynamics due to prioritizing local safety. MARL has been adapted to address the specific complexities of social navigation. A primary distinction from standard MARL---which typically focuses on robot-robot coordination and fast goal reaching---is the additional requirement to model asymmetric human-robot interactions and maintain social awareness. SAMARL \cite{Wang2023RL} achieves this by utilizing a hybrid spatial-temporal transformer to capture and align high-order dependencies in both human-robot and robot-robot interactions. Such methods are computationally efficient during inference, but developing a reward function that truly emulates human-like behavior remains an active research area.

In contrast to these existing works, our method uses a rolling-horizon diffusion model. Our framework leverages learned human motion priors directly from data, naturally denoising paths that are socially aware. This generative foundation, combined with multi-agent coordination techniques from MAPF, enables our system to maintain high social acceptance and coordination for up to 15 robots, exceeding the scale of existing learning-based benchmarks.

\section{Preliminaries}
We now introduce the problem definition and background on diffusion models for multi-robot path planning.
\subsection{Problem Definition}
Our task is multi-robot navigation in a 2D workspace $\mathcal{W} \subset \mathbb{R}^2$ containing $N$ robots $\{X_i\}_{i=1}^N$, $M$ humans  $\{Y_j\}_{j=1}^M$, and $Q$ static obstacles $\{O_k\}_{k=1}^Q$ where each $O_k \subseteq \mathcal{W}$.  Time is discretized into uniform timesteps separated by $\Delta t$ (set to 0.1 seconds in our experiments). We assume access to a perfect prediction function that provides $\mathcal{H}_j = \{Y_j^{0}, Y_j^{1}, \cdots, Y_j^{H - 1}\}$, the predicted path of human $Y_j$ over the next $H$ timesteps, 
where $H$ is the planning horizon (set to 32 in our experiments), and $Y_j^t \in \mathcal{W}$ is the position of human $Y_j$ $t$ timesteps after the prediction is made. Each robot $X_i$ is assigned a start and goal position $s_i, g_i \in \mathcal{W}$. A full path of robot $X_i$ is written as a time-indexed sequence of waypoints $\tau_i = \{\tau^1_i, \tau^2_i, \cdots, \tau^{L_i}_i\}$ with $\tau^1_i = s_i$ and $\tau^{L_i}_i = g_i$, where $L_i$ is referred to as the length of the path, and $\tau^t_i \in \mathcal{W}$ is the position of robot  $X_i$ at time $t$. 
Our objective is to find a set of full paths $\{\tau_i\}_{i=1}^N$ that do not collide with each other (robot-robot collision), with any humans (robot-human collision), or with any obstacles (robot-obstacle collision).

We seek to learn a set of human-like paths $\{\tau_i\}_{i=1}^N$. While hand-crafted cost functions struggle to capture complex social dynamics, we instead learn directly from a dataset of human-like movement. We also seek to minimize the maximum path completion time, $\max(L_i)$, across all robots. 

\subsection{Diffusion Models and Constraints}
Diffusion models learn to emulate complex patterns from a dataset through an iterative denoising process. In this framework, motion planning is formulated as a probabilistic inference problem where the model learns a prior over expert paths. The generated paths should also satisfy task objectives, such as avoiding obstacles. The goal is to maximize the posterior probability of a path $\tau_i$ given task objectives $\mathcal{O}$:
\begin{equation}
    \arg \max_{\tau_i} \log p(\tau_i|\mathcal{O}) = \arg \min_{\tau_i} (\mathcal{J}(\tau_i) - \log p(\tau_i)),
    \label{eq:objective}
\end{equation}
where $\mathcal{J}(\tau_i)$ is a cost associated with the task objectives, and $\log p(\tau _i)$ represents the prior to the learned data distribution.

The training phase uses a denoising diffusion probabilistic model \cite{ddpm} to approximate the distribution of expert paths. This involves a forward diffusion process that gradually adds noise to an expert path $^0\tau_i$ (of fixed length) through an $S$-step sequence $^0\tau_i, ^1\tau_i, \cdots, ^S\tau_i$. As $S$ becomes sufficiently large, this process is designed to reach a stationary state where the original structure of the path is erased, leaving only Gaussian noise. Therefore, to generate a new path $^0\tau_i$, we sample a noisy path $^S\tau_i  \sim \mathcal{N}(0, I)$ from a Gaussian distribution of the same length as the expert path and train a temporal U-net $\epsilon_{\theta}$ to iteratively reverse the process by predicting and removing noise $\epsilon$ added at each step $s$. 

During inference, to ensure that generated paths satisfy the task objectives $\mathcal{O}$, classifier guidance gradients are included in the sampling loop. Following MMD~\cite{shaoul2025multirobot}, the denoising is a probabilistic transition sampled from the distribution
\begin{equation}
    ^{s-1}\tau_i \sim \mathcal{N}(\mu_{\theta}(^{s}\tau_i, s)+ \eta\beta_{s-1}\nabla_{\tau _ i}\mathcal{J}(^s\tau_i), \beta_{s-1}I),
    \label{eq:denoise}
\end{equation}
where $\mu_{\theta}(^{s}\tau_i, s)$ represents the reverse transition mean predicted by the temporal U-net $\epsilon_{\theta}$, $\eta$ is a scalar that controls the weight of the guidance, and $\beta_{s-1}$ is the variance of the denoising schedule. The collective ``guidance" $\nabla_{\tau_i}\mathcal{J}(^s\tau_i)$ is the gradient of the task objectives from Equation \ref{eq:objective}. One way to incorporate all task objectives is to use a weighted sum $\mathcal{J} = \sum_{o=1}^O\lambda_o\mathcal{J}_o$, where each weighted cost optimizes for a different task objective in $\mathcal{O}$. After each denoising step, the first and last waypoints are fixed to the start and goal, ensuring that the model outputs valid full paths.

In multi-robot path planning, the generated paths need to be free of robot-robot collisions. MMD accomplishes this by using a high-level CBS algorithm to find collisions and generate robot-robot constraints, and a low-level diffusion-based planner to generate individual paths that avoid constrained regions. These constraints are included in the guidance term $\mathcal{J}$ from Equation \ref{eq:denoise} as a cost term $\mathcal{J}_{\mathrm{c}}$. Robot-robot constraints are represented as spheres in the workspace, and paths with waypoints inside the spheres incur higher costs, resulting in gradient steps pushing the generation outside of the constrained regions. Specifically, we represent a sphere constraint for robot $X_i$ as $\langle X_i, p, t \rangle$, where $p \in \mathcal{W}$ is the center of the sphere, and $t$ is the associated timestep. The corresponding cost function is formulated as 
\begin{equation}
    \mathcal{J}_{\mathrm{c}}({}^{s}\tau_{i}) := \sum_{u = t-t_0}^{t+t_0} \max (r - d ( {}^{s}\tau_{i}^{u}, p) , 0 ), \label{eq:constraint}
\end{equation}
where $t_0$ specifies the half-length of the constrained time window, $r$ is the fixed radius of the sphere, and $d( {}^{s}\tau_{i}^{u}, p)$ is the distance of robot $X_i$ to $p$ at time $u$ along path $^{s}\tau_{i}$. 

A significant limitation of current diffusion-based path planners, including MMD and CoBL-Diffusion, is that all generated paths contain a fixed number of waypoints. Because consecutive waypoints are separated by a constant time interval $\Delta t$, the robot's arrival time is fixed. This ``fixed-time problem" introduces critical performance issues. In particular, the robot's velocity is dictated primarily by its distance to the goal, leading to dangerously high velocities for distant goals or unnaturally slow, stuttering movements for nearby goals. This behavior is problematic in human-shared environments, where maintaining an acceptable pace is vital. Furthermore, the fixed-time problem prevents diffusion-based planners from optimizing execution time.

\section{Method}
We introduce MRRD, a 3-level diffusion-based path planning algorithm for multi-robot navigation in human-shared environments. At the top level, our method plans robot paths iteratively using a rolling-horizon scheme~\cite{li2021lifelongmultiagentpathfinding}, interleaving planning and execution to form a set of full long-horizon paths. In the middle level, we resolve robot-robot collisions within horizons and assign constraints for path generation. At the bottom level, we generate human-like paths for each robot from a single-robot diffusion policy learned from human path data.

\begin{algorithm}[t]
\caption{MRRD: Rolling-Horizon Planner}
\label{alg:continuous_plan}
\begin{algorithmic}[1]
\Require Start positions $\{s_i\}_{i=1}^{N}$, goal positions $\{g_i\}_{i=1}^{N}$, planning horizon $H$, execution horizon $K$


\While{$\exists i: ||g_i - s_i|| > \delta$} \label{alg:loop}
    \vspace{0.5mm}
    \State $\{\mathcal{H}_j\}_{j=1}^M \gets \textsc{GetHumanPaths}(H)$ \label{alg:gethumanpaths}
    \State $\{\pi_i\}_{i=1}^{N} \gets \textsc{PlanSegment}(\{s_i\}, \{g_i\}, \{\mathcal{H}_j\})$ \label{plansegment}
    \State $\{s_i\}_{i=1}^N \gets \textsc{Execute}(\{\pi _i\}$, $K$) \label{alg:execute}
\EndWhile
\end{algorithmic}
\end{algorithm}

\subsection{Rolling-Horizon Planning}
Algorithm \ref{alg:continuous_plan} shows our top-level rolling-horizon planner. Unlike prior diffusion-based planners, which plan the full path from start to finish, MRRD generates short path segments in a rolling-horizon fashion. Since we do not generate full robot paths directly, the human path prediction function only needs to look ahead to the planning horizon $H$ (line \ref{alg:gethumanpaths}). In line \ref{plansegment}, we generate path segments $\pi_i$ of length $H$ for all robots. During the execution step (line \ref{alg:execute}), we execute $K < H$ steps of this plan, where $K$ is called the execution horizon. The positions of the robots after execution are set as their new start positions for the next iteration. This process is repeated (line \ref{alg:loop}) until all robots have reached their goals within a certain tolerance $\delta$ (set to $0.2$ meters in our experiments). In practice, we cap the maximum number of iterations (set to 11 in our experiments) of the while loop such that the robot execution ends.

\begin{algorithm}[t]
\caption{Multi-Robot Planner (\textsc{PlanSegment})}
\label{alg:plan_rolling_horizon_parallel}
\begin{algorithmic}[1]
\Require Start positions $\{s_i\}_{i=1}^{N}$, goal positions $\{g_i\}_{i=1}^{N}$, human paths $\{\mathcal{H}_j\}_{j=1}^M$
\Ensure Collision-free paths $\{\pi_i\}_{i=1}^{N}$
\Statex \hrulefill
\Statex \textbf{Phase 1: Segment Generation}
\State $\textit{root} \gets \textsc{Node}()$
\State $\textit{root}.C_i \gets \emptyset \hspace{0.05in}\forall i\in \{1\cdots N\}$ 
\State $\{\textit{root}.\pi_i\}_{i=1}^{N} \gets \textsc{DiffPlanParallel}(\{s_i\}, \{g_i\}, \{\mathcal{H}_j\})$ \label{alg:plan_parallel}
\State $\textit{root}.\textit{colls} \gets \textsc{GetCollisions}(\textit{root})$ \label{alg:get_colls}
\State $\textsc{Open} \gets \{\textit{root}\}$ \label{alg:open_list}

\Statex \hrulefill
\Statex \textbf{Phase 2: Collision Resolution}

\While{$\textsc{Open} \neq \emptyset$}
    \State Sort nodes in $\textsc{Open}$ by number of collisions
    \State $P \gets \textsc{Open}.\textsc{Pop}()$
    \If{$P.\textit{colls} = \emptyset$\label{alg:goal_test}}
        \State \Return $\{P.\pi_i\}_{i=1}^N$
    \EndIf
    \State $p, t, X_i, X_j \gets \textsc{GetConstraint}(P.\textit{colls}[0])$ \label{alg:get_constraint}
    
    \For{$k \in \{i, j\}$} 
        \State $P' \gets \textsc{Copy}(P)$
        \State $P'.C_k \gets P'.C_k \cup \{\langle X_k, p, t\rangle\}$ \label{alg:add_constraint}
        \State $P'.\pi_k \gets \textsc{DiffPlan}(s_{k}, g_{k}, \{\mathcal{H}_j\}, P'.C_k)$ \label{alg:diffplan}
        \State $P'.\textit{colls} \gets \textsc{GetCollisions}(P')$
        \State $\textsc{Open} \gets \textsc{Open} \cup \{P'\}$
    \EndFor
\EndWhile
\State \Return $\emptyset$
\end{algorithmic}
\end{algorithm}

\subsection{Multi-Robot Coordination}
\begin{figure}[t]
  \centering
  \includegraphics[scale=0.46]{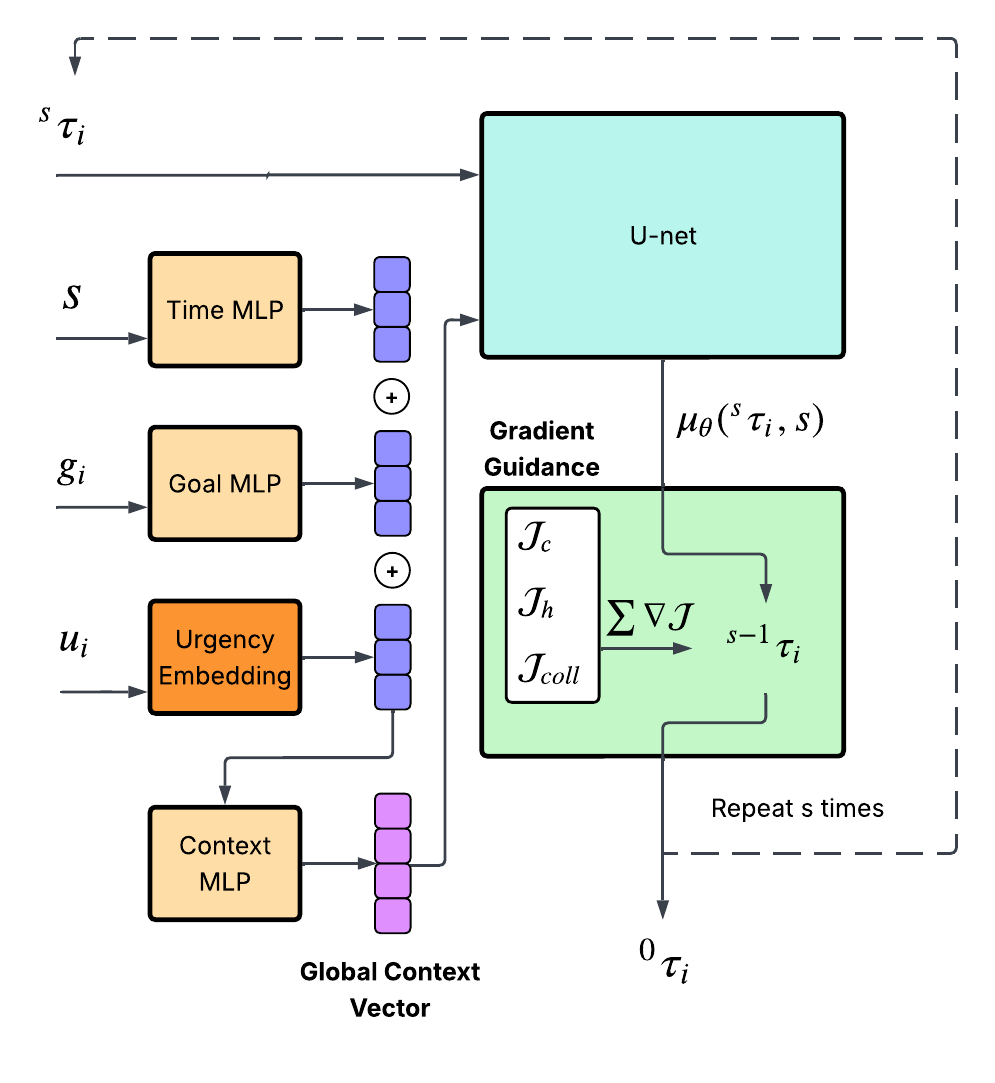}
  \caption{Architecture overview of our diffusion model. The denoising step, robot goal, and urgency are provided as context to a U-net for denoising. Then, task objectives of constraint avoidance, human avoidance, and obstacle avoidance are applied as gradient guidance. This pattern is repeated for $S$ iterations, resulting in a fully denoised path $^0\tau_i$.}
  \label{model}
\end{figure}
MRRD treats each rolling horizon as its own localized multi-robot planning problem, as shown in Algorithm \ref{alg:plan_rolling_horizon_parallel}. This level is inspired by MMD \cite{shaoul2025multirobot} and is meant to resolve robot-robot collisions. The process requires several key steps. 

\textbf{Segment Generation}: The path segment generation process begins with the creation of a root node. Each node contains $\{\pi _i\}_{i=1}^N$, a set of path segments of length $H$ for each robot, $\{C _i\}_{i=1}^N$, a set of constraint sets imposed on each robot (initially empty), and \textit{colls}, a list of robot-robot collisions between the path segments. During each planning horizon, we start by calling the diffusion-based planner to plan a path segment for each robot (line \ref{alg:plan_parallel}). 
The returned path segments are guaranteed to be free of robot-human and robot-obstacle collisions. 
If no such path segments are generated, we terminate the algorithm and report failure (not shown in \Cref{alg:plan_rolling_horizon_parallel}). Then, we find the list of all (robot-robot) collisions between path segments in line \ref{alg:get_colls} and add the root node to the OPEN list in line \ref{alg:open_list}.

Since diffusion models are known to be computationally intensive, we implement parallelized path generation in line \ref{alg:plan_parallel} to greatly increase inference speed. Unlike MMD, which creates its root node by generating the path segment of each robot sequentially, we parallelize the process by generating all path segments in one call. This change allows the system to scale to a larger number of robots while maintaining the high-frequency updates required for real-time navigation.

\textbf{Collision Resolution}: If the path segments generated in the root node contain collisions, MRRD initiates a tree search process to resolve them. In each iteration, the node $P$ with the fewest collisions is popped from the OPEN list for expansion. Upon identifying a collision between two robots $X_i$ and $X_j$ at position $p$ and time $t$ (line \ref{alg:get_constraint}), the search ``splits" the problem by generating two child nodes. Each child node  $P'$ inherits the existing constraints of the parent, but in line \ref{alg:add_constraint}, we add a new sphere constraint $\langle X_k, p, t \rangle$ with $k \in \{i,j\}$, preventing robot $X_k$ from occupying a sphere centered at position $p$ around time $t$. Each child replans its respective robot's path segment using the newly added constraint, along with all existing constraints, generating a new path segment guided around the constrained regions (line \ref{alg:diffplan}). 
The resulting child nodes are then evaluated for new collisions and added back to the OPEN list. This process continues until a node containing zero collisions is retrieved (line~\ref{alg:goal_test}), at which point the set of path segments is returned as a valid solution for the current planning horizon.

\subsection{Single-Robot Diffusion}
\label{sec:single_robot_diffusion}
The bottom level of MRRD is the diffusion-based planner that generates a single-robot path segment of length $H$ while taking into account human social norms, obstacle avoidance, and the sphere constraints imposed by \Cref{alg:plan_rolling_horizon_parallel}. The primary innovation in this level is the transition from rigid endpoint-pinned conditioning to more flexible temporal goal conditioning. Unlike previous approaches such as Motion Planning Diffusion (MPD) \cite{carvalho2023mpd} and MMD, which typically perform global planning by fixing both the first and last waypoints to the start and goal positions at each denosing step, MRRD fixes only the first waypoint while allowing the last waypoint to remain flexible. Furthermore, MRRD conditions the diffusion model on the goal position $g_i$ and an urgency term $u_i$, where $u_i$ represents the total number of waypoints from the start to the goal. Intuitively, an urgency value $u_i$ implies that the current generated path segment advances the agent by approximately $H/u_i$ of the remaining distance toward the goal. 
Below, we provide more details on how this modification changes our model architecture, denoising, and inference. 

\subsubsection{Model Architecture and Training}
Our architecture is based on the one utilized in MPD. Like MPD, we take a noisy path $^S\tau_i \sim \mathcal{N}(0, I)$ of length $H$ and denoise it to generate $^0\tau_i$. As shown in Fig. \ref{model}, the goal $g_i$ and the urgency term $u_i$ are new inputs to the model in addition to the random path and the denoising step $s$. We obtain a goal embedding using an MLP and an urgency embedding using an embedding dictionary. These inputs are concatenated and fed through a context layer to form the global context vector, providing key information to the model about the goal and desired speed.

Our diffusion-based planner is trained on human walking data, providing a set of paths $\{\psi_i\}$ of length $T > H$. For each path $\psi_i = \{\psi_i^1, \psi_i^2, \cdots, \psi_i^T\}$, we sample an integer $u_i$ uniformly from $[H, T)$. Then, we use the first $H$ waypoints of $\psi_i$ as the expert path and set the corresponding goal input to $g_i = \psi_i^{u_i}$, effectively training the model to learn the urgency term $u_i$ as a ``time-to-goal." %

\subsubsection{Denoising and Guidance}
Our diffusion-based planner is called in both line \ref{alg:plan_parallel} and line \ref{alg:diffplan} of \Cref{alg:plan_rolling_horizon_parallel}, where the inputs $\{\mathcal{H}_j\}$ and $P'.C_k$ are used in the guidance steps of the diffusion process. Specifically, the following guidance terms are summed to obtain our complete guidance at each denoising step, $\mathcal{J} = \sum_{o=1}^O\lambda_o\mathcal{J}_o$.

\textbf{Constraint Costs}: As in MMD, we penalize robots that violate the sphere constraints in $P'.C_k$. This guidance term is the same as it is defined in Equation \ref{eq:constraint}.

\textbf{Human Distance Costs}: We also penalize robots for being in range of human personal space. We take inspiration from CoBL-Diffusion \cite{MizutaLeung2024} in designing the cost. The human distance cost at time $t$ for human $Y_j$ is modeled as
\begin{equation*}
P_j(t) = \max (0, R_{out} - D) + \max (0, R_{in} - D )^2,
\end{equation*}
where $R_{in}$ and $R_{out}$ (set to 0.1 meters and 0.7 meters in our experiments) specify a core radius with squared penalty and a close radius with linear penalty, and $D = d(^s\tau_i^t, Y_j^t)$ is the distance between robot $X_i$ and human $Y_j$ at time $t$. To efficiently manage dense crowds, for each time $t$, we identify the $m$ (set to 3 in our experiments) humans with the highest penalties $P_j(t)$ and sort them in descending order: $P_{(1)}(t) \ge P_{(2)}(t) \ge \dots \ge P_{(m)}(t)$. The final cost $\mathcal{J}_{h}(^s\tau_i)$ is the rank-weighted sum of these top $m$ penalties over the planning horizon $H$:
\begin{equation*}
\mathcal{J}_{h}(^s\tau_i) = \sum_{t=0}^{H-1} \sum_{j=1}^{m} (m - j + 1) P_{(j)}(t).
\end{equation*}

\textbf{Obstacle Costs}: This cost (similar to the one in MPD \cite{carvalho2023mpd}) evaluates how deeply each waypoint of the generated path penetrates into static obstacles, using the environment's Signed Distance Field (SDF). The SDF returns negative values for points inside obstacles and positive values for those outside. The cost is weighted by a precision factor $\kappa = 1/\sigma_{coll}^2$, where $\sigma_{coll}$ (set to 1.0 in our experiments) controls the penalty strength. The per-timestep obstacle error is the clamped SDF value (with $M_{\textit{coll}} = 0.02$ in the experiments):
\begin{equation*}
    e(t) = \max(M_{\textit{coll}},\; \text{SDF}(^s\tau^t_i)).
\end{equation*}
The total collision cost over the path is:
\begin{equation*}
    \mathcal{J}_{\textit{coll}}(^s\tau_i) = \kappa\sum_{t=0}^{H - 1} e(t).
\end{equation*}

After the guidance is applied, we set the first waypoint $^s\tau^0_i$ of the generated path to $s_i$, so that our final path $^0\tau_i$ starts from the current position of the robot.

\subsubsection{Model Inference}
During inference, the diffusion model is queried in batches. We vary $u_i$ linearly from a range $[u_{\textit{min}}, u_{\textit{max}})$ to generate a batch of $B$ path segments with varying speeds. This diversity ensures that the planner has multiple options, such as moving quickly through a transient opening or slowing down to wait for a pedestrian to pass. In our experiments, $u_{\textit{max}}$ is set to $200$, and we use the following equation to give $u_{\textit{min}}$ a larger value when the robot is far from its goal to prevent rushing and slowly relax it as the robot gets closer:
\begin{equation*}
u_{\textit{min}} =
\begin{cases}
\lambda\cdot||s_i - g_i||, & \text{if } ||s_i - g_i|| < \delta_{\text{close}} \\
\lambda\delta_{\text{close}}, & \text{otherwise.}
\end{cases}
\end{equation*}
where $\delta_{\text{close}}$ is set to 4 and $\lambda$ is set to 10.

Then, among the generated paths, we first discard infeasible ones, namely those that lead to robot-human or robot-obstacle collisions, and then select the path with the smallest $u_i$, corresponding to the highest speed. 
By doing so, we resolve the fixed-time problem faced by earlier models, allowing the diffusion process to adapt the path's velocity profile within the fixed $H$-step window.



\section{Experimental Analysis}

We evaluate the performance of MRRD in maps with various human crowds, numbers of robots, and static obstacles. Experiments are performed on a machine with an AMD Threadripper 3990X 64-core processor and an NVIDIA RTX 3090Ti. Our implementation is in Python and is based on the official code for MMD \cite{shaoul2025multirobot}. 

\subsection{Training Setup}
The diffusion model is trained on 20,000 20-second paths, discretized into $T = 200$ waypoints from the ETH human pedestrian dataset \cite{pellegrini2009you} (making $\Delta t = 0.1$ seconds). Paths are augmented by randomizing their starting location and rotation to provide the model with a variety of human-like movements. Our model sets the planning horizon to $H = 32$ steps and the execution horizon to $K = 16$ steps, which correspond to 3.2 seconds and 1.6 seconds, respectively.

\subsection{Baseline Setups}
We compare our method to baselines that capture social awareness: CoBL-Diffusion \cite{MizutaLeung2024}, a diffusion-based planner, and Navi-STAR \cite{wang2023navistar}, an RL-based planner with preference learning. Since the SAMARL repository was unavailable, we used Navi-STAR, an earlier single-robot planner from the same group. We also include an ablation planner of MRRD that generates initial paths sequentially, without parallelization. CoBL-Diffusion and Navi-STAR, originally single-robot planners, were adapted for multi-robot planning using a prioritized-planning approach, where paths are planned sequentially to avoid collisions with higher-priority robots.

\begin{figure}[t]
  \centering
  \includegraphics[scale=0.52]{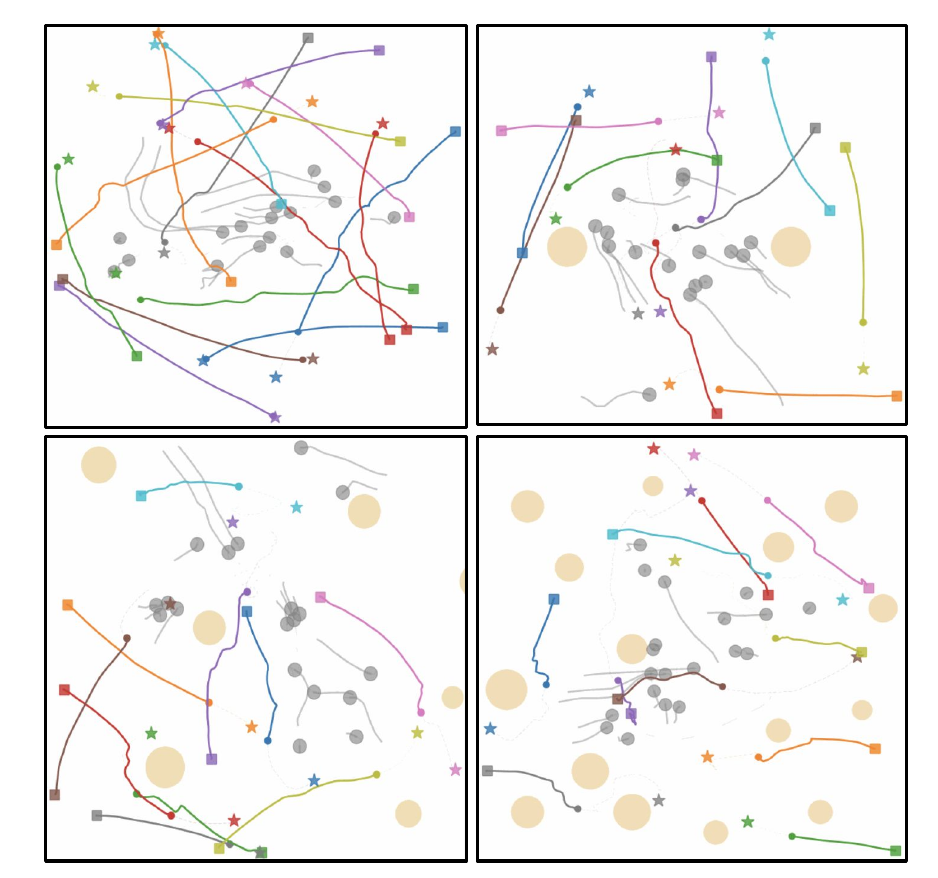}
  \caption{Snapshot of ongoing runs. Robots (colored circles) move from starts (squares) to goals (stars). Humans are light gray; tan circles indicate obstacles. Layouts clockwise from top-left: Empty, Simple, Spheres, and Spheres Sparse.}
  \label{fig:obstacles}
\end{figure}

\subsection{Crowds in Empty Map}
\begin{table*}[t]
\vspace{1em}
\centering
\caption{Comparison of Collision Rates and Goal Scores.}
\label{tab:safety_results}
\setlength{\tabcolsep}{5.3pt} 
\begin{tabular}{l rrrr rrrr rrrr rrrr}
\toprule
& \multicolumn{4}{c}{6 Robots} & \multicolumn{4}{c}{9 Robots} & \multicolumn{4}{c}{12 Robots} & \multicolumn{4}{c}{15 Robots} \\
\cmidrule(lr){2-5} \cmidrule(lr){6-9} \cmidrule(lr){10-13} \cmidrule(lr){14-17}
Method & R-R$\downarrow$ & R-H$\downarrow$ & SR$\uparrow$ & GR$\downarrow$ & R-R & R-H & SR & GR & R-R & R-H & SR & GR & R-R & R-H & SR & GR \\
\midrule
CoBL-Diffusion & .10 & .22 & .68 & 1.00 & .10 & .34 & .56 & 1.80 & .24 & .38 & .44 & 1.28 & .28 & .54 & .32 & 1.46 \\
Navi-STAR         & .38 & .76 & .16 & .90 & .58 & .85 & .04 & 1.00 & .90 & 1.00 & .00 & .92 & .98 & .96 & .00 & 1.12 \\
Ours (No Parallelization)  & \textbf{.00} & .04 & .96 & .36 & \textbf{.00} & \textbf{.04} & \textbf{.96} & .36 & \textbf{.00} & .12 & .88 & .40 & \textbf{.00} & .14 & .86 & .44 \\
\textbf{Ours}  & \textbf{.00} & \textbf{.00} & \textbf{1.0} & \textbf{.29} & \textbf{.00} & \textbf{.04} & \textbf{.96} & \textbf{.24} & \textbf{.00} & \textbf{.08} & \textbf{.92} & \textbf{.34} & \textbf{.00} & \textbf{.08} & \textbf{.92} & \textbf{.37} \\
\bottomrule
\addlinespace
\multicolumn{17}{l}{\small \textit{Note: R-R: Robot-Robot Collision Rate, R-H: Robot-Human Collision Rate, GS: Goal-Reaching ($m$), SR: Success Rate}.}
\end{tabular}
\end{table*}

\begin{table*}[t]
\label{table:t2}
\centering
\caption{Comparison of Time and Acceleration Profiles.}
\label{tab:efficiency_results}
\begin{tabular}{l rrr rrr rrr rrr}
\toprule
& \multicolumn{3}{c}{6 Robots} & \multicolumn{3}{c}{9 Robots} & \multicolumn{3}{c}{12 Robots} & \multicolumn{3}{c}{15 Robots} \\
\cmidrule(lr){2-4} \cmidrule(lr){5-7} \cmidrule(lr){8-10} \cmidrule(lr){11-13}
Method & $T_P$$\downarrow$ & $T_T$$\downarrow$ & Acc$\downarrow$ & $T_P$ & $T_T$ & Acc & $T_P$ & $T_T$ & Acc & $T_P$ & $T_T$ & Acc \\
\midrule
CoBL-Diffusion & 25.2 & \textbf{8.0} & .242 & 71.7 & \textbf{8.0} & .402 & 49.2 & \textbf{8.0} & .316 & 60.2 & \textbf{8.0} & .319 \\
Navi-STAR         & 7.1 & 19.1 & .193 & 10.1 & 20.0 & .123 & 13.4 & 20.0 & .207 & 17.3 & 20.0 & .179 \\
Ours (No Parallelization)  & 14.5 & 17.0 & \textbf{.043} & 34.6 & 16.8 & \textbf{.047} & 33.3 & 17.6 & \textbf{.046} & 38.6 & 17.6 & .047 \\
\textbf{Ours}  & \textbf{2.9} & 16.3 & .046 & \textbf{3.3} & 16.0 & .051 & \textbf{4.3} & 17.4 & \textbf{.046} & \textbf{8.8} & 17.6 & \textbf{.046} \\
\bottomrule
\addlinespace
\multicolumn{13}{l}{\small \textit{Note: $T_P$: Planning Time (s), $T_T$: Travel Time (s), Acc: Mean Acceleration ($m/s^2$).}}
\end{tabular}
\end{table*}

\begin{table*}[t]
\centering
\caption{Comparison of Collision Rates and Goal Scores (Obstacles).}
\label{tab:safety_5maps}
\footnotesize
\setlength{\tabcolsep}{3.5pt} 
\begin{tabular}{l rrrrr rrrrr rrrrr rrrrr}
\toprule
& \multicolumn{5}{c}{Empty} & \multicolumn{5}{c}{Simple} & \multicolumn{5}{c}{Spheres Sparse} & \multicolumn{5}{c}{Spheres} \\
\cmidrule(lr){2-6} \cmidrule(lr){7-11} \cmidrule(lr){12-16} \cmidrule(lr){17-21}
Method & R-R$\downarrow$ & R-H$\downarrow$ & R-O$\downarrow$ & SR$\uparrow$ & GR$\downarrow$ & R-R & R-H & R-O & SR & GR & R-R & R-H & R-O & SR & GR & R-R & R-H & R-O & SR & GR \\
\midrule
CoBL-Diffusion & .04 & .70 & \textbf{.00} & .30 & 1.20 & .16 & .42 & .20 & .42 & 1.09 & .10 & .64 & .48 & .18 &  1.56 & .26 & .60 & .94 & .00 & 2.08 \\
\textbf{Ours} & \textbf{.00} & \textbf{.04} & \textbf{.00} & \textbf{.96} & \textbf{.50} & \textbf{.00} & \textbf{.10} & \textbf{.18} & \textbf{.72} & \textbf{.42} & \textbf{.00} & \textbf{.02} & \textbf{.12} & \textbf{.88} & \textbf{.54} & \textbf{.00} & \textbf{.10} & \textbf{.30} & \textbf{.56} & \textbf{.52} \\
\bottomrule
\addlinespace
\multicolumn{13}{l}{\small \textit{Note: R-O: Robot-Obstacle Collision Rate.}}
\end{tabular}
\end{table*}

\begin{table*}[t]
\centering
\caption{Comparison of Time and Acceleration Profiles (Obstacles).}
\label{tab:efficiency_5maps}
\footnotesize
\begin{tabular}{l rrr rrr rrr rrr}
\toprule
& \multicolumn{3}{c}{Empty} & \multicolumn{3}{c}{Simple} & \multicolumn{3}{c}{Spheres Sparse} & \multicolumn{3}{c}{Spheres} \\
\cmidrule(lr){2-4} \cmidrule(lr){5-7} \cmidrule(lr){8-10} \cmidrule(lr){11-13}
Method & $T_P$$\downarrow$ & $T_T$$\downarrow$ & Acc$\downarrow$ & $T_P$ & $T_T$ & Acc & $T_P$ & $T_T$ & Acc & $T_P$ & $T_T$ & Acc \\
\midrule
CoBL-Diffusion & 52.3 & \textbf{8.0} & .313 & 49.6 & \textbf{8.0} & .325 & 55.3 & \textbf{8.0} & .348 & 54.1 & \textbf{8.0} & .363 \\
\textbf{Ours} & \textbf{3.7} & 17.6 & \textbf{.049} & \textbf{3.5} & 17.6 & \textbf{.051} & \textbf{4.3} & 17.6 & \textbf{.054} & \textbf{5.1} & 17.4 & \textbf{.065} \\
\bottomrule
\end{tabular}
\end{table*}

We use a 20 m $\times$ 20 m empty map with 10-20 human crowds from the UCY pedestrian dataset \cite{lerner2007crowds}. We tested 50 trials in 6, 9, 12, and 15 robots with randomly initialized start and goal positions (7-12 meters apart) and initial separation from humans (over 1 meter). We evaluated the generated paths using 6 metrics: The \textit{robot-robot collision rate} (R-R) and the \textit{robot-human collision rate} (R-H) are the fractions of trials that experience a robot-robot collision and a robot-human collision, respectively; \textit{goal reaching} (GR) is the average distance from each robot to its goal at the end of the algorithm; the \textit{success rate} is the fraction of trials that experience no collisions of any type; the \textit{planning time} ($T_P$) is the overall planning time (summed over all rolling horizons for ours); the \textit{travel time} ($T_T$) is the execution time for the robot paths; and the \textit{average acceleration} (Acc) is a measure of smoothness. The upper left square in Fig. \ref{fig:obstacles} shows an example run of our algorithm in progress.

Tables \ref{tab:safety_results} and \ref{tab:efficiency_results} show the performance of the four algorithms. MRRD achieves zero robot-robot collisions across all test cases (6 to 15 robots), validating the effectiveness of the multi-robot planner. In contrast, Navi-STAR’s R-R collision rate scales poorly, reaching 0.98 in the 15-robot scenario. Even CoBL-Diffusion still suffers from collision rates of up to 0.28. MRRD only halts due to an imminent R-H collision at a rate of 0.08, while the baselines' R-H collision rates range from 0.22 to 1.00. This demonstrates that MRRD does not sacrifice human safety for multi-robot coordination.

The robustness of MRRD is most evident in the success rate metric. Although MRRD maintains a high success rate of 0.92 even with 15 robots, the baselines do not scale. CoBL-Diffusion's success rate drops to 0.32, and Navi-STAR fails to complete a single successful trial at the 15-robot scale. While these methods are individually very successful at navigating single robots through dense human crowds, they were not directly designed to handle a multi-robot scenario efficiently. If any of the generated paths results in a collision, the trial is automatically deemed a failure. Furthermore, the data confirms that the ``fixed-time" constraint inherent in CoBL-Diffusion's architecture is a critical flaw. Because CoBL-Diffusion is forced to generate a path within a rigid temporal window, it cannot adapt to various crowds, distances, and robot-robot interactions. This results in higher R-R/R-O collision rates and failure to reach goals, evidenced by a 1.46-meter goal-reaching score. MRRD offers a much better balance: it reaches goals more accurately (GR of 0.37 meters) and produces significantly smoother trajectories, with acceleration profiles that are 5-8x lower than CoBL-Diffusion's erratic movements. This property is evident in simulated visualizations, where MRRD exhibits smooth and human-like movement.

\begin{figure}[t]
  \centering
  \includegraphics[scale=0.24]{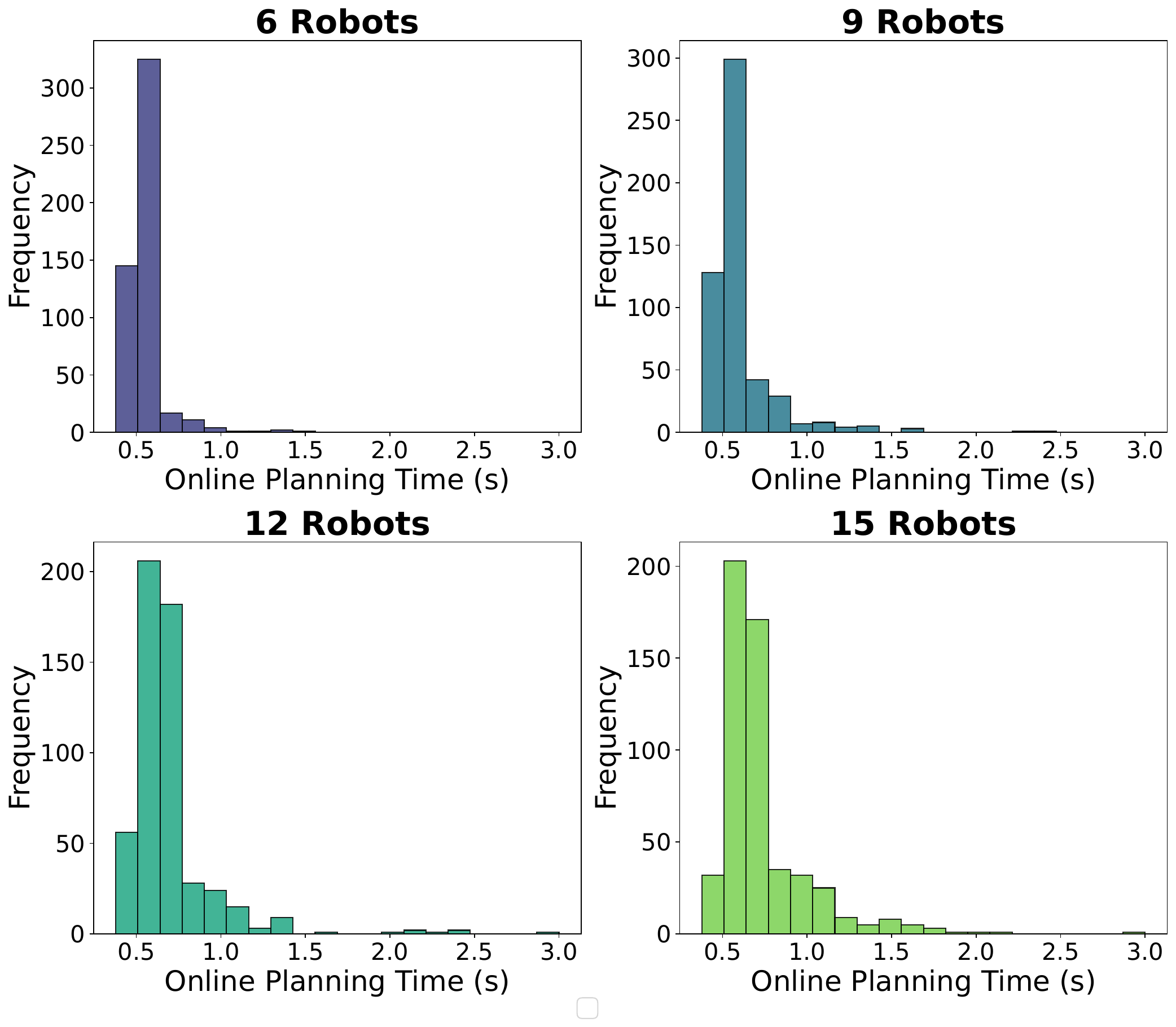}
  \caption{Histogram of online planning times of MRRD for 4 different robot counts. The x-axis is the time it takes for Algorithm \ref{alg:plan_rolling_horizon_parallel} to plan path segments for all robots, and the y-axis is the count. Planning times must remain below the execution time of 1.6 seconds for each horizon.}
  \label{fig:rh_times}
\end{figure}

MRRD also demonstrates the best planning efficiency. At the 15-robot scale, MRRD’s planning time is only 8.8 seconds, making it 7 times faster than CoBL-Diffusion (60.2 seconds), 2 times faster than Navi-STAR (17.3 seconds), and 4 times faster than our MRRD without parallelization (38.6 seconds). The histograms in Fig. \ref{fig:rh_times} illustrate the computational efficiency of MRRD across varying robot densities. They display ``planning times," representing the time taken for each call of \textsc{PlanSegment}. The distributions exhibit a primary peak centered between 0.5 seconds and 0.75 seconds. The larger tails found in the 12-robot and 15-robot charts correspond to scenarios that require extensive CBS splitting and multiple diffusion replanning calls to resolve inter-robot collisions. Even at the scale of 15 robots, the vast majority of planning times at each horizon successfully conclude within the execution horizon of 1.6 seconds.

\subsection{Crowds in Obstacle Maps}
Next, we test our algorithms on more complex scenarios with static obstacles. Although neither baseline natively supports planning in environments with obstacles, we provided a competitive comparison by augmenting CoBL-Diffusion with an obstacle-avoidance guidance term. We introduce a new metric \textit{robot-obstacle collision rate}, which is the fraction of trials with a robot-obstacle collision. We set the number of robots to 10 and ran 50 trials on the empty map and the three 20 m $\times$ 20 m obstacle maps, shown in Fig. \ref{fig:obstacles}. 

Tables \ref{tab:safety_5maps} and \ref{tab:efficiency_5maps} present the results. Even as obstacle density increases, MRRD maintains a perfect 0.00 R-R collision rate across all tested scenarios, whereas CoBL-Diffusion reaches a high of 0.26. MRRD also effectively navigates robots around humans, keeping the R-H collision rate at or below 0.12 even in crowded obstacle fields---a 6x improvement over the baseline in the Spheres map. Even in the densest scenario, MRRD’s R-O rate (0.30) is less than a third of CoBL-Diffusion’s (0.94). While CoBL-Diffusion’s success rate reduces from 0.18 to 0 when moving from the Spheres Sparse to Spheres maps, MRRD remains resilient, maintaining strong SRs of 0.88 and 0.56, respectively. This performance gap highlights that MRRD is fundamentally more capable of navigating high-density spaces where traditional diffusion-based planners often fail. Table \ref{tab:efficiency_5maps} reveals that MRRD achieves a 10 to 15 times decrease in planning time, solving complex obstacle fields in just 5.1 seconds compared to CoBL-Diffusion’s 54.1 seconds. Furthermore, MRRD follows the trend of producing significantly more stable trajectories than CoBL-Diffusion, with mean acceleration profiles that are roughly 6 times lower.

\section{Conclusion}
In this paper, we introduced a 3-level diffusion-based algorithm MRRD for real-time, multi-robot path planning in human-crowded environments. MRRD integrates a rolling-horizon planner and a multi-robot coordinator with a diffusion model trained on human paths. By generating short, dynamically adaptable path segments rather than long, hard-conditioned paths, our method enables greater dynamic planning than recent diffusion-based works. Our experimental results quantitatively show fewer collisions and greater goal-reaching capability than other baselines trained on human data and input. Qualitatively, the paths are smooth and socially aware with human speed and fluidity. Future work includes adding invariance to the denoising process for path compliance \cite{xiao2025safe} and integrating human movement prediction modules directly into the diffusion denoising step. 

\section*{Acknowledgments}
This work was partially supported by the National Science Foundation under grant numbers \#$2328671$ and \#$2441629$. 

\addtolength{\textheight}{-12cm}   

\bibliographystyle{ieeetr}
\bibliography{citation}

\end{document}